\documentclass[11pt,a4paper]{article}
\usepackage[hyperref]{acl2018}
\usepackage{times}
\usepackage{latexsym}

\usepackage{url}
\usepackage[utf8x]{vietnam}
\usepackage[utf8x]{inputenc}
\usepackage{graphicx}

\AtBeginDocument{}
\AtBeginDocument{}
\AtBeginDocument{}

\aclfinalcopy 

\title{A Feature-Based Model for Nested Named-Entity Recognition at VLSP-2018 NER Evaluation Campaign}

\author{Pham Quang Nhat Minh\\
Alt Vietnam Co., Ltd\\
92 Trieu Viet Vuong, Hai Ba Trung, Hanoi\\
  {\tt pham.minh@alt.ai}
}

\date{}

\begin{document}
\maketitle

\begin{abstract}
In this report, we describe our participant named-entity recognition system at VLSP 2018 evaluation campaign. We formalized the task as a sequence labeling problem using BIO encoding scheme. We applied a feature-based model which  combines word, word-shape features, Brown-cluster-based features, and word-embedding-based features. We compare several methods to deal with nested entities in the dataset. We showed that combining tags of entities at all levels for training a sequence labeling model (joint-tag model) improved the accuracy of nested named-entity recognition.
\end{abstract}

\section{Introduction}
\label{sec:intro}

\begin{table*}[!t]
\centering
\begin{tabular}{|l|ccc|}
\hline 
\textbf{Word} & \textbf{Level-1 Tag} & \textbf{Level-2 Tag} & \textbf{Joint Tag}\\
 \hline
ông & O & O & O+O\\ 
Ngô\_Văn\_Quý & B-PER & O & B-PER+O\\
- & O & O & O+O\\
Phó & O & O & O+O\\
Chủ\_tịch & O & O & O+O\\
UBND & O & B-ORG & O+B-ORG\\
TP & B-LOC & I-ORG & B-LOC+I-ORG\\
Hà\_Nội & I-LOC & I-ORG & I-LOC+I-ORG\\
 \hline
\end{tabular}
\caption{\label{tbl:jointTag} Generating joint-tags by combing entity tags at all levels of a token}
\end{table*}

Named-entity recognition (NER) is an important task in information extraction. The task is to identify in a text, spans that are entities and classify them into pre-defined categories. There have been some conferences and shared tasks for evaluating NER systems in English and other languages, such as MUC-6~\cite{Sundheim1995OverviewOR}, CoNLL 2002~\cite{Sang2002IntroductionTT} and CoNLL 2003~\cite{Sang2003IntroductionTT}.

In Vietnamese language, VLSP 2016 NER evaluation~\cite{Huyen2016} is the first evaluation campaign that aims to systematically compare NER systems for Vietnamese language. Similar to CoNLL 2003 shared-task, in VLSP 2016, four named-entity types were considered: person (PER), organization (ORG), location (LOC), and miscellaneous entities (MISC). In VLSP 2016, organizers provided the training/test with gold word segmentation, PoS and chunking tags. While that setting can help participant teams to reduce effort of data processing and solely focus on developing NER algorithms, it is not so realistic setting. In VLSP 2018 NER evaluation, only raw texts with XML tags were provided. Therefore, we need to choose appropriate Vietnamese NLP tools for preprocessing steps such as word segmentation, PoS tagging, and chunking.

In the report, we describe our NER system at VLSP 2018 NER evaluation campaign. We applied a feature-based model which combines word, word-shape features, Brown-cluster-based features, and word-embedding-based features and adopted Conditional Random Fields (CRF)~\cite{Lafferty:2001} for training and testing.

In the VLSP 2018 NER task, similar as VLSP 2016, there are nested entities the NER dataset. An entity may contain other entities inside them. We categorize entities in VLSP 2018 NER dataset into three levels.

\begin{itemize}
\item Level-1 entities are entities that do not contain other entities inside them. For example: <ENAMEX TYPE=``LOC''>Hà Nội</ENAMEX>.
\item \begin{flushleft}Level-2 entities are entities contain only level-1 entities inside them. For example: <ENAMEX TYPE=``ORG''>UBND thành phố <ENAMEX TYPE=``LOC''>Hà Nội</ENAMEX></ENAMEX>.\end{flushleft}
\item \begin{flushleft}Level-3 entities are entities that contain at least one level-2 entity and may contain some level-1 entities. For example <ENAMEX TYPE=``ORG''>Khoa Toán, <ENAMEX TYPE=``ORG''>ĐHQG <ENAMEX TYPE=``LOC''>Hà Nội</ENAMEX></ENAMEX></ENAMEX>\end{flushleft}
\end{itemize}

In our data statistics, we see that the number of level-3 entities is too small compared with the number of level-1 and level-2 entities, so we decided to ignore them in building the model. We just consider level-1 and level-2 entities.

In order to deal with nested named-entities, we investigated two methods. The first method trains separated models for each level of entities. The second method trains a single model on the training data in which tags are generated by combing entity tags of entities of all levels. Table~\ref{tbl:jointTag} shows an example of how we combined entity tags at all levels of a token to create join tags.

We showed that combining tags of entities at all levels for training a sequence labeling model (joint-tag model) improved the accuracy of nested named-entity recognition.

The rest of the paper is organized as follows. In section~\ref{sec:system}, we described our participant NER system. In section~\ref{sec:eval}, we present our evaluation results. Finally, section~\ref{sec:conclusion} gives conclusions about the work.

\section{System description}
\label{sec:system}

We formalize NER task as a sequence labeling problem by using the B-I-O tagging scheme and we apply a popular sequence labeling model, Conditional Random Fields to the problem. In this section, first we present how we preprocess the data and then present features that we used in our model.

\subsection{Preprocessing}

In our NER system, we performed sentence and word segmentation on the data. For sentence segmentation, we just used a simple regular expression to detect sentence boundaries that match the pattern: period followed by a space and upper-case character. Actually, to produce result submissions, we also try not to perform sentence segmentation.

For word segmentation, we adopted RDRsegmenter~\cite{NguyenNVDJ2018} which is the state-of-the-art Vietnamese word segmentation tool. Both training and development data are the converted into data files in CoNLL 2003 format with two columns: words and their BIO tags. Due to errors of word segmentation tool, there may be boundary-conflict problem between entity boundary and word boundary. In such cases, we decided to tag words as ``O'' (outside entity).

\subsection{Features}

Basically, features in the proposed NER model are categorized into word, word-shape features, features based on word representations including word clusters and word embedding. Note that, we extract unigram and bigram features within the context surrounding the current token with the window size of $5$. More specifically, for a feature $F$ of the current word, unigram and bigram features are as follows.

\begin{itemize}
\item \textbf{unigrams}: $F$[-2], $F$[-1], $F$[0], $F$[1], $F$[2]
\item \textbf{bigrams}: $F$[-2]$F$[-1], $F$[-1]$F$[0], $F$[0]$F$[1], $F$[1]$F$[2]
\end{itemize}

\subsubsection{Word Features}
	
We extract word-identity unigrams and bigrams within the window of size 5. We use both word surfaces and their lower-case forms. Beside words, we also extract prefixes and suffixes of surfaces of words within the context of the current word. In our model, we use prefixes and suffixes of lengths from 1 to 4 characters.

\subsubsection{Word Shapes}

In addition to word identities, we use word shapes to improve prediction ability, especially for unknown or rare words and reduce data spareness problem. We used the same word shapes as presented in~\cite{1803.04375}.

\subsubsection{Brown cluster-based features}

Brown clustering algorithm is a hierarchical clustering algorithm for assigning words to clusters~\cite{Brown:1992:CNG:176313.176316}. Each cluster contains words which are semantically similar. Output clusters are represented as bit-strings. Brown-cluster-based features in our NER model include whole bit-string representations of words and their prefixes of lengths 2, 4, 6, 8, 10, 12, 16, 20. Note that, we only extract unigrams for Brown-cluster-based features.

In experiments, we used the Brown clustering implementation of Liang~\cite{liang2005semi} and applied the tool on the raw text data collected through a Vietnamese news portal. We performed word clustering on the same preprocessed text data which were used to generate word embeddings in~\cite{le2017empirical}. The number of word clusters used in our experiments is 5120.

\subsubsection{Word embeddings}

Word-embedding features have been used for a CRF-based Vietnamese NER model in~\cite{le2017empirical}. The basic idea is adding unigram features corresponding to dimensions of word representation vectors. 

In the paper, we apply the same word-embedding features as in~\cite{le2017empirical}. We generated pre-trained word vectors by applying Glove~\cite{pennington2014glove} on the same text data used to run Brown clustering. The dimension of word vectors in 25.

\section{Evaluation}
\label{sec:eval}

\subsection{Data sets}

\begin{table*}[!t]
\centering
\begin{tabular}{| l | ccc | ccc | ccc |}
\hline
\textbf{Type} & \multicolumn{3}{|c|}{\textbf{Train}} & \multicolumn{3}{|c|}{\textbf{Dev}} & \multicolumn{3}{|c|}{\textbf{Test}}\\
\hline
& \textbf{Level-1} & \textbf{Level-2} & \textbf{Level-3} & \textbf{Level-1} & \textbf{Level-2} & \textbf{Level-3} & \textbf{Level-1} & \textbf{Level-2} & \textbf{Level-3}\\
\hline
LOC & 8831 & 7 & 0 & 3043 & 2 & 0 & 2525 & 2 & 0\\
ORG & 3471 & 1655 &  63 & 1203 & 690 & 14 & 1616 & 557 & 22\\
PER & 6427 & 0 & 0 & 2168 & 0 & 0 & 3518 & 1 & 0\\
MISC & 805 & 1 & 0 & 179 & 1 & 0 & 296 & 0 & 0\\
\hline
Total & 19534 & 1663 & 63 & 6593 & 694 & 14 & 7955 & 561 & 22\\
 \hline
\end{tabular}
\caption{\label{tbl:data} Number of entities of each type in each level in train/dev and test set}
\end{table*}

Table~\ref{tbl:data} showed the data statistics on training set, development set, and official test set. The number of organization entities (ORG) at level 3 is too small, so we only consider level-1 and level-2 entities in training and evaluation. Level-2 entities are almost of ORG types.

\subsection{Evaluation Measures}

We used Precision, Recall, F1 score as evaluation measures. Note that, due to the fact that word segmentation may cause boundary conflict between entities and words, we convert words in the data into syllables before we evaluate Precision, Recall, F1 scores.

We consider four entity types: LOCATION, MISCELLANEOUS, ORGANIZATION, and PERSON in evaluation, and use the evaluation script of CoNLL-2013 for evaluation.

\subsection{NER models}

For evaluation on the development set, we train three NER models as follows on the training data of VLSP 2018 NER task.

\begin{itemize}
\item Level-1 model is trained by using level-1 entity tags.
\item Level-2 model is trained by using level-2 entity tags.
\item Joint model is trained using joint tags which combine level-1 and level-2 tags of each word.
\end{itemize}

\subsection{Results}

\begin{table}[!t]
\centering
\begin{tabular}{|l|ccc|}
\hline 
\textbf{Model} & \textbf{Precision} & \textbf{Recall} & \textbf{F1}\\
 \hline
 Level-1 Model & 91.04 & 84.41 & 87.6\\
 Joint Model & 90.42 & 84.72 & 87.47\\
 \hline
\end{tabular}
\caption{\label{tbl:dev1} Evaluation results on dev set of recognizing level-1 entities}
\end{table}

\begin{table}[!t]
\centering
\begin{tabular}{|l|ccc|}
\hline
\textbf{Method} & \textbf{Precision} & \textbf{Recall} & \textbf{F1}\\
 \hline
 Level-2 & 85.81 & 72.44 & 78.56\\
 Joint Model & 84.36 & 77.06 & 80.54\\
 \hline
\end{tabular}
\caption{\label{tbl:dev2} Evaluation results on dev set of recognizing level-2 entities}
\end{table}

Table~\ref{tbl:dev1} and Table~\ref{tbl:dev2} shows the evaluation results on development set of recognizing level-1 and level-2 entities, respectively. The level-1 model obtained slightly better F1 score than joint model in recognizing level-1 entities while joint model outperformed level-2 model in recognizing level-2 entities. We also see that the level-2 model got higher precision than joint model but much lower recall than joint model. A plausible explanation for that phenomena is that information of level-1 tags helps to recognize more level-2 entities.

\subsection{Result Submissions}

We trained models on the data set obtained by combining provided training and development data and used the trained models for recognizing entities on the test set.

In order to produce submitted results, we use methods as follows.

\begin{itemize}
\item Using level-1 and level-2 model for recognizing level-1 and level-2 entities, respectively. We refer this method as \textbf{Separated} method.
\item We use joint model to recognize joint tags for each word of a sentence, then split joint tags into level-1 and level-2 tags. We refer this method as \textbf{Joint} method.
\item We use the joint model for recognizing level-2 entities and level-1 model for recognizing level-1 entities. We refer this method as \textbf{Hybrid} method.
\end{itemize}

In recognition, there are some cases that predicted level-1 entities contains level-2 entities inside them. In such cases, we omit predicted level-2 entities inside predicted level-1 entities. The reason is that accuracy of level-1 entity recognition on dev set is much higher than the accuracy of level-2 entity recognition.

We submitted six runs at VLSP 2018 NER evaluation campaign as showed in Table~\ref{tbl:submission}. We try two preprocessing approaches: with sentence segmentation and without sentence segmentation. The reason why we try those preprocessing approaches is that we would like to know the influence of sequence lengths on the accuracy of our model.

\begin{table}[!t]
\centering
\begin{tabular}{| l | lc |}
\hline
\textbf{Runs} & \textbf{Method} & \textbf{Sent Segmentation}\\
 \hline
Run-1 & Hybrid & YES\\
Run-2 & Hybrid & NO\\
Run-3 & Joint & YES\\
Run-4 & Joint & NO \\
Run-5 & Separated & YES\\
Run-6 & Separated & NO\\
 \hline
\end{tabular}
\caption{\label{tbl:submission} Six submitted runs}
\end{table}

\begin{table}[!t]
\centering
\begin{tabular}{| l | ccc |}
\hline
\textbf{Run} & \textbf{Precision} & \textbf{Recall} & \textbf{F1}\\
 \hline
Run-1 & 76.08 & 70.68 & 73.28\\ 
Run-2 & 76.75 & 70.37 & 73.42\\ 
Run-3 & 76.32 & 70.25 & 73.16\\ 
Run-4 & 76.16 & 70.98 & \textbf{73.48}\\ 
Run-5 & 75.70 & 70.28 & 72.89\\ 
Run-6 & 76.26 & 69.90 & 72.94\\ 
 \hline
\end{tabular}
\caption{\label{tbl:NestedResult}  Official evaluation results  on test set, which consider entities at all levels}
\end{table}

\begin{table}[!t]
\centering
\begin{tabular}{| l | ccc |}
\hline
\textbf{Category} & \textbf{Precision} & \textbf{Recall} & \textbf{F1}\\
 \hline
PER & 79.30 &  79.68 & 79.49\\
LOC & 79.21 & 79.69 & 79.45\\
ORG & 66.83 & 60.17 & 63.33\\
MISC & 51.40 & 25.00 & 33.64\\
All & 76.16 & 70.98 & 73.48\\
 \hline
\end{tabular}
\caption{\label{tbl:detailedResult}  Evaluation results  of Run-4 on test set for each entity category}
\end{table}

Table~\ref{tbl:NestedResult} shows the official evaluation results for our six submitted runs. As indicated in the table, run 4 which uses \textbf{Joint} model obtained the highest F1 score among six runs. Using \textbf{Joint} model or \textbf{Hybrid} model obtained better F1 scores than using \textbf{Separated} methods. We also see that the difference between a system that  performs sentence segmentation and a system that does not perform sentence segmentation is very small.

Table~\ref{tbl:detailedResult} shows the Precision, Recall, F1 scores for each entity category of run 4.

\begin{table}[!t]
\centering
\begin{tabular}{| l | ccc |}
\hline
\textbf{Run} & \textbf{Precision} & \textbf{Recall} & \textbf{F1}\\
 \hline
Run-1 & 73.82 & 79.43 & 76.52\\
Run-2 & 73.45 & 80.04 & 76.60\\ 
Run-3 & 73.21 & 79.56 & 76.26\\ 
Run-4 & 73.95 & 79.33 & 76.55\\
Run-5 & 73.80 & 79.46 & 76.53\\
Run-6 & 73.46 & 80.08 &\textbf{76.63}\\
 \hline
\end{tabular}
\caption{\label{tbl:officialEval1}  Evaluation results  on test set for level-1 entities}
\end{table}

\begin{table}[!t]
\centering
\begin{tabular}{| l | ccc |}
\hline
\textbf{Run} & \textbf{Precision} & \textbf{Recall} & \textbf{F1}\\
 \hline
Run-1 & 43.24 & 82.94 & 56.84\\
Run-2 & 43.06 & 82.59 & 56.61\\
Run-3 & 45.20 & 81.41 & \textbf{58.12}\\
Run-4 & 44.48 & 82.51 & 57.80\\
Run-5 & 39.32 & 83.08 &  53.38\\
Run-6 & 36.83 & 84.15 & 51.24\\
 \hline
\end{tabular}
\caption{\label{tbl:officialEval2}  Evaluation results  on test set for level-2 entities}
\end{table}

Table~\ref{tbl:officialEval1} and Table~\ref{tbl:officialEval2} showed the evaluation results on test set of six submitted runs for level-1 and level-2 entities, respectively. 

Run-6 (using level-1 and level-2 models separately without sentence segmentation) obtained the best accuracy of recognizing level-1 entities among submitted runs ($76.63$\%) and Run-3 (Joint model, sentence segmentation) obtained the best accuracy of recognizing level-2 entities ($58.12\%$).

Using joint model obtained better F1 scores of recognizing both levels of entities than just those of the model trained on solely on level-1 and level-2 entity tags. That result is consistent with the result on the development set.



\section{Conclusions}
\label{sec:conclusion}

We haved presented a feature-based model for Vietnamese named-entity recognition and evaluation results at VLSP 2018 NER evaluation campaign. We compared several methods for recognizing nested entities. Experimental results showed that combining tags of entities at all levels for training a sequence labeling model improved the accuracy of nested named-entity recognition. As the future work, we plan to investigate deep learning methods such as BiLSTM-CNN-CRF~\cite{ma2016end} for nested named entity recognition.

\bibliography{VLSP2018}
\bibliographystyle{acl_natbib}

\end{document}